# SPARK: Spatial Policy-driven Adaptive Reinforcement learning for Knowledge distillation

M.J.Aashik Rasool
aashikrasool@gachon.ac.kr
Shabir Ahmad
shabir@gachon.ac.kr
Gisung Oh
colin@collaborators.com
Teagkeun Whangbo
tkwhangbo@gachon.ac.kr

Department of IT Convergence Engineering, Gachon University, Sujeong-gu, Seongnam-si, Gyeonggi-do 461-701, Republic of Korea

**Abstract**

Low-bit quantization enables deployment of image restoration (IR) networks on resource-constrained devices, but introduces rounding noise that disproportionately degrades high-frequency regions such as edges and fine textures. Existing knowledge distillation (KD) methods apply distillation signals uniformly across all spatial locations, overlooking the varying reconstruction difficulty across image regions. To address this, we propose SPARK (Spatial Policy-driven Adaptive Reinforcement learning for Knowledge distillation), a framework that adaptively allocates distillation effort using a lightweight reinforcement learning (RL) policy network. At each training step, a difficulty feature extractor computes four signals, namely Laplacian variance, pixel variance, student reconstruction error, and teacher-student knowledge gap, which are fed into a compact policy CNN that produces a stochastic spatial weight map to modulate the KD loss during quantization-aware training (QAT). SPARK is IR task-agnostic, adds no inference cost, and integrates into any existing QAT pipeline without architectural changes. Experiments on benchmark datasets demonstrate that SPARK consistently outperforms PTQ, QAT, and state-of-the-art (SOTA) KD approaches across multiple student architectures, achieving reconstruction quality closest to the full-precision teacher under significant computational constrain.

## 1 Introduction

Deep neural networks have emerged as the dominant technique in current computer vision, capable of performing a wide range of low-level vision tasks such as super-resolution (SR), denoising, dehazing, low-light enhancement (LLE), and general IR [12, 14, 23, 24]. Contemporary approaches regularly create visually realistic reconstructions; however, that advancement has come at a high computational cost [10, 12]. SOTA IR networks require billions of multiply accumulate operations per pixel, which renders them challenging to deploy on smartphones, security cameras, drones, and embedded vision systems, where on-device processing is becoming more widespread [10, 12]. Low-bit quantization, which uses 8 or fewer bits, provides a principled solution by replacing floating-point weights and activations with compact integer representations, resulting in significant memory, latency, and energy savings[4, 13].

However, quantization reduces the quality of output in IR tasks more than in high-level vision tasks such as classification, object detection, or image segmentation[4]. Pixel-regression objectives are highly sensitive to the rounding noise injected by low-bit operators, and the absence of a softmax-style smoothing stage means that small numerical errors propagate directly to the output [13]. Quantized models often failed to produce visually better output relative to their full-precision counterparts [13], with the largest drops appearing on high-frequency content such as edges, fine textures, repetitive structures, and detail-rich regions, which are precisely the regions human observers attend to most and where enhancement quality is judged.

On the other hand, KD is a well-known approach for training lightweight models under the supervision of complex teacher models[3, 18, 23]. Here, a teacher model supervises the student model so that the student's outputs match the teacher's outputs [3]. Existing KD formulations for IR, however, apply the distillation signal uniformly across the spatial domain, averaging a single loss over every pixel of the image [3]. This treatment is at odds with what IR networks actually need to learn. Smooth, low-frequency regions such as flat skies, uniform backgrounds, or homogeneous surfaces are easy for both the teacher and the student, so forcing the student to mimic the teacher there contributes little useful gradient and can crowd out learning in harder regions. Edges, textures, and detail-rich patches, by contrast, occupy a small fraction of pixels but account for a disproportionate share of the perceptual error, and they are exactly where quantization noise is most damaging[13]. A loss that weights all locations equally is, in effect, a loss that underweights the locations that matter.

Even though there are several hand-crafted approaches introduced by researchers, such as edge masks[25], gradient-based weighting, or fixed difficulty heuristics have been explored [18, 23], each approach commits in advance to a single notion of where to focus and remains static throughout training. We argue that the right spatial allocation of distillation effort is itself a learnable, state-dependent decision. It depends on the current capacity of the student, the residual gap to the teacher, the local image statistics, and the characteristics of the specific degradation being addressed, all of which evolve as training progresses. We therefore introduce spatial KD weighting as a sequential decision problem and learn it with a lightweight RL policy.

Concretely, we introduce a Spatial KD Policy that, at every training step, observes a four-channel difficulty extractor of the current sample, namely Laplacian variance (edges), pixel variance (texture), the student's reconstruction error against the ground truth, and the teacher–student knowledge gap, and emits a stochastic spatial weight map. This map spatially weights an MSE distillation loss between the quantized student and the frozen full-precision teacher, while a standard L1 reconstruction loss continues to anchor the student to the ground truth (GT). Crucially, the sampled action is detached before being applied to the KD loss, so the policy is trained by reward signals tied to reconstruction quality rather than by pixel-wise gradients. This formulation enables the policy network to explore allocation strategies that are inaccessible to purely differentiable objectives. The policy itself is tiny, consisting of three convolutional layers operating on an $8 \times 8$ grid. It adds no cost at inference, since it is discarded after training, and plugs into any existing QAT pipeline for IR without architectural changes to the student. The comparison between the conventional KD approach and our proposed SPARK is visually illustrated in Fig. 1 . The SPARK framework is IR task-agnostic and applies to any IR task with paired training data, including LLE, denoising, and SR.

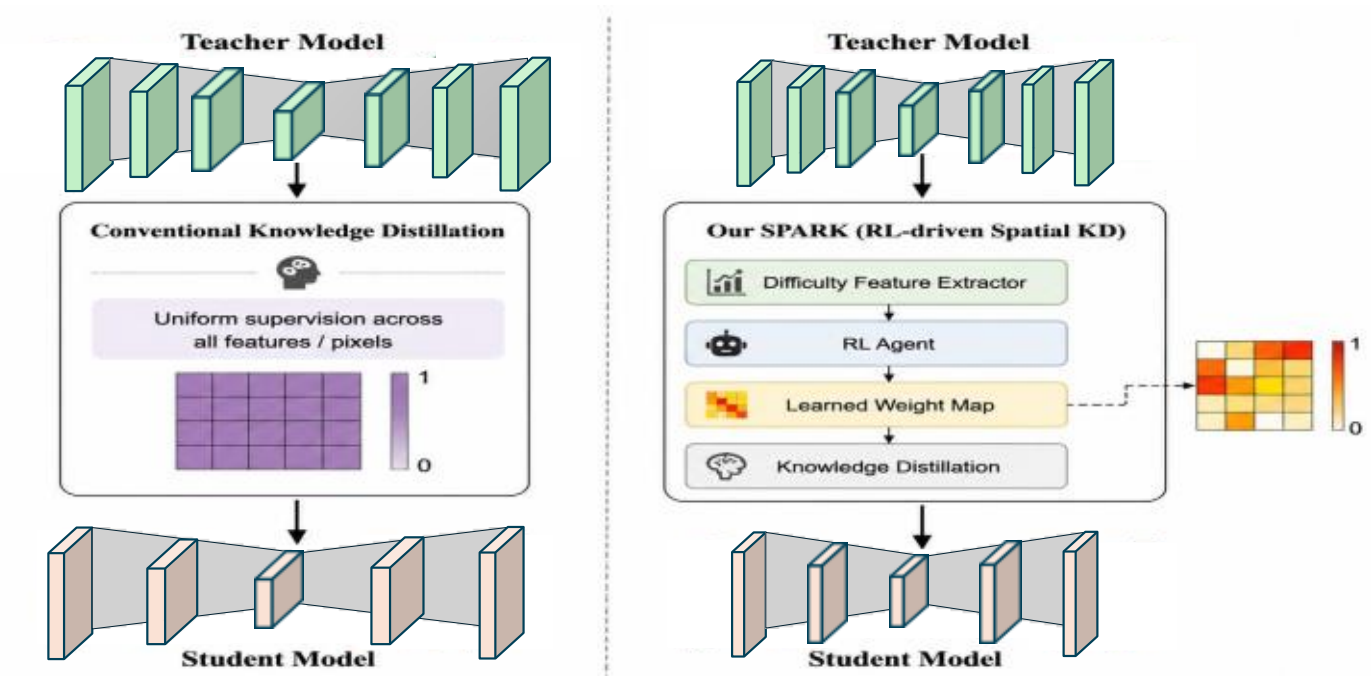

Figure 1: Comparison between conventional KD and the proposed SPARK framework. Conventional KD (left) applies a uniform supervision signal across all spatial regions, whereas SPARK (right) employs an RL-driven policy that generates a spatially adaptive weight map to focus distillation pressure on challenging regions such as edges and fine textures.

In summary, our main contributions are as follows:

- We identify the spatial uniformity of existing KD losses as a key bottleneck for low-bit IR tasks and motivate spatially adaptive distillation as the missing ingredient for quantization-robust restoration.
- We propose SPARK, an RL-driven Spatial KD Policy that consumes interpretable difficulty features (edge, texture, student error, teacher-student gap) and produces a per-region weight map learned end-to-end during QAT. The policy emits a stochastic Gaussian-sampled action with detached gradients, which decouples policy learning from the supervised reconstruction objective and enables exploration of non-trivial spatial allocations. To the best of our knowledge, this is the first RL-based spatial weighting strategy for KD in low-bit IR, and it is broadly applicable across diverse tasks such as LLE, denoising, and SR.
- We conducted extensive experiments on standard benchmarks across multiple IR tasks and student architectures, demonstrating that our proposed SPARK consistently outperforms uniform KD baselines, with the most significant gains concentrated in edge and texture-rich regions that correspond to the known failure modes of low-bit IR.

# 2 Related Works

## 2.1 Knowledge Distillation for Low-Level Vision tasks

In IR tasks, KD approaches are utilized to transfer knowledge from the teacher network to a lightweight student network, guiding it to learn effectively. KD for IR has been extensively explored in recent literature. He et al. introduced FAKD [18], which aligns the spatial affinity matrices derived from the feature maps of the teacher and student networks. Building on this line of work, Li et al. proposed MiPKD [8] to mitigate the semantic gap between teacher and student features, performing feature mixing and stochastic network block mixing in the latent space for SR. Zhou et al. introduced DCKD [24], which employs dynamic contrastive regularization to adaptively adjust the distilled solution space based on the student's learning state. A distribution mapping module is additionally used to align pixel-level category distributions between the teacher and student, addressing the limitations of fixed solution spaces and L1-only objectives in previous IR distillation methods. Despite their effectiveness, existing KD methods for IR treat all spatial regions equally and lack a principled mechanism to adaptively allocate distillation effort to harder regions, a limitation that becomes particularly critical under the constrained capacity of low-bit QAT[9].

### 2.2 Reinforcement Learning in Low-Level Vision tasks

RL has been explored as a flexible mechanism for introducing decision-making and adaptive behavior into vision systems where deterministic supervision is insufficient[7]. In high-level vision, RL has been applied to neural architecture search[16], attention, active object localization, and early-exit policies for efficient inference[7]. For low-level vision, Yu et al. proposed RL-Restore [17], which learns an agent to sequentially select restoration operations from a predefined toolbox to handle compound, unknown degradations. ReLLIE [22] reformulates LLE as a sequential curve-estimation problem and trains a lightweight policy with non-reference rewards, enabling customized, image-specific enhancement. However, its potential for guiding distillation itself, in particular, learning where and how strongly a student should mimic its teacher across spatial regions, remains largely unexplored, especially under the constrained capacity of low-bit QAT, which motivates our proposed SPARK framework.

## 3 Method

In our SPARK framework [Fig.2], a compact, quantized student network is trained to replicate the reconstruction quality of a larger, pretrained teacher network. Rather than applying uniform distillation across all regions, our method dynamically assigns relevance weights to distinct image regions based on their reconstruction complexity, using an RL policy network.

### 3.1 Teacher Network

The teacher network is a large-capability IR model that has been pre-trained to convergence and frozen throughout training (Fig. 2). It take LR inputs of shape (B, 3, H, W) and generates a high-quality output $I_t$ of shape (B, 3, H, W), which serves as both a reconstruction reference and a soft supervisory signal to the student.

### 3.2 Student Network

The student network is a lightweight model or self-quantized model of the teacher architecture trained using QAT that is intended for deployment in resource-constrained contexts[3].

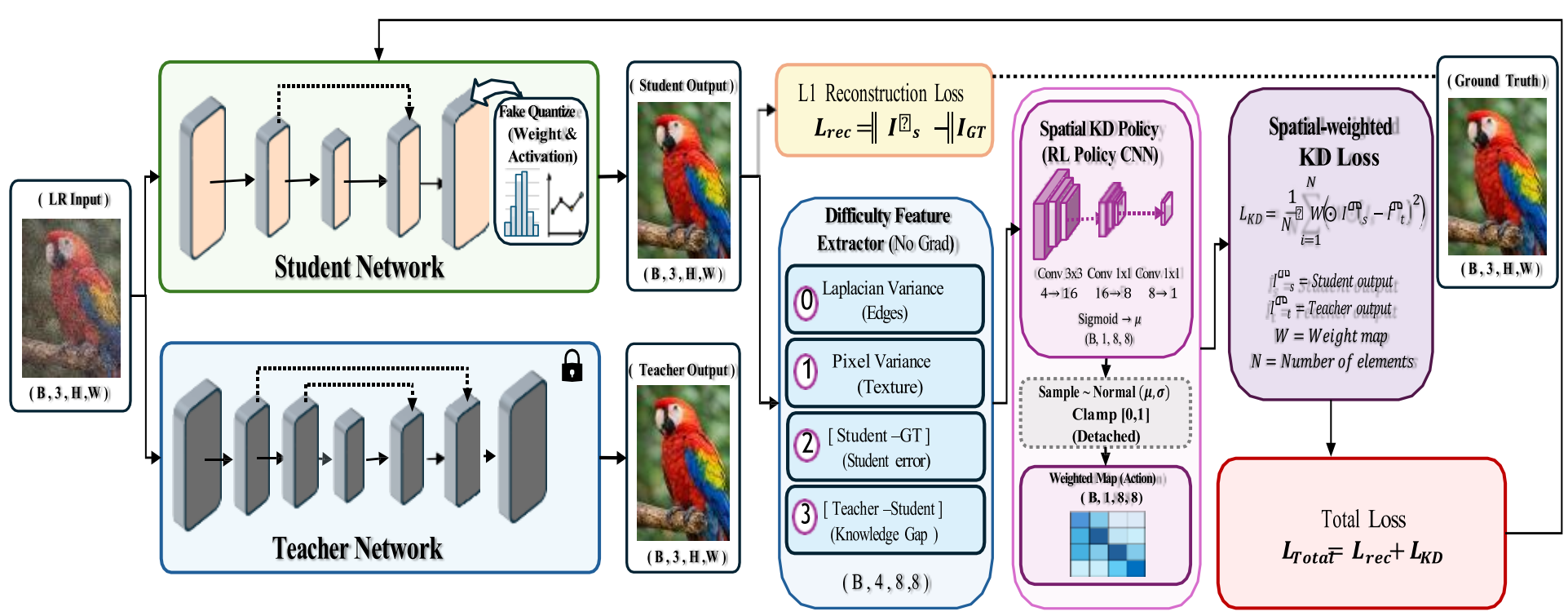

Figure 2: Overview of the proposed SPARK framework, illustrating the student and teacher network pipeline, the difficulty feature extractor, and the spatial KD policy via a RL policy CNN that generates a spatial weight map for computing the spatially weighted KD loss, combined with the reconstruction loss to form the total training objective.

During training, false quantization is applied to both weights and activations to mimic deployment time quantization behavior in the forward pass, while full-precision gradients are kept for optimization. The student processes the same LR input and produces output $I_s$ of shape (B, 3, H, W).

## 3.3 Difficulty Feature Extractor

To capture spatial reconstruction difficulty, we extract four complementary feature maps from both networks' representations, with no gradients flowing through this module to ensure it functions purely as an analytical signal. The first feature, Laplacian Variance, assesses edge sharpness, capturing high-frequency regions where strong edges make reconstruction intrinsically difficult. The second feature, Pixel Variance, measures local texture complexity, as highly textured areas require more representational capacity from the student. The third feature, $|Student - GT|$, represents the student loss, represents the student loss map, which quantifies the student's per-pixel reconstruction loss relative to the ground truth.

## 3.4 The Spatial Policy Network

The difficulty features are fed into an RL Policy Network, a lightweight CNN. Policy-based RL optimizes a parameterized policy $\pi_\theta$ , rendering it suitable for continuous action spaces such as spatial weight map prediction, which are high-dimensional and non-discrete. This differs from value-based RL approaches.

Policy gradient methods are better suited for learning smooth, spatially diverse weight distributions than value-based approaches. This policy network predicts a spatial weight map $W$ that reflects the relative importance of each region in the distillation process. Specifically, policy-based RL offers several key advantages in this context. It allows for direct optimization of a stochastic strategy over continuous weight values, eliminating the instability caused by discretizing spatial importance scores. The inherent stochasticity of the policy facilitates systematic exploration throughout the training process, enabling the network to identify non-trivial spatial weight configurations that deterministic, purely supervised objectives would otherwise be incapable of capturing.

The output is processed through a Sample $\sim \mathcal{N}(\mu, \sigma)$ layer and clamped to $[0, 1]$ to produce a normalized, differentiable weight map of shape $(B, 1, 8, 8)$. The random sampling technique from a learnt Gaussian distribution promotes exploration during training, avoiding the policy from falling into a rigid deterministic weighting method and maintaining strong generalization across various image contents.

## 3.5 Reward Function

The reward function is critical for training the RL policy network because it offers a scalar feedback signal that guides the policy to generate spatial weight maps that increase student reconstruction quality.

Considering the policy network utilizes a spatially adaptive weight map instead of utilizing a constant loss, the reward must indicate the reconstruction quality of the student's output at each training step. In each training iteration, the reward $r$ is determined as the improvement in reconstruction quality between the current student output $\hat{I}_s$ and the ground truth $I_{GT}$, assessed using a combination of PSNR and SSIM:

$$r = \alpha \cdot \Delta\mathrm{PSNR}(\hat{I}_s, I_{GT}) + \beta \cdot \Delta\mathrm{SSIM}(\hat{I}_s, I_{GT}) \tag{1}$$

In here, $\Delta$PSNR and $\Delta$SSIM represent the change in reconstruction quality relative to a student output learns without spatial weighting. $\alpha$ and $\beta$ are scalar weighting factors that balance the two indicators. A positive reward indicates that the spatial weight map created by the policy resulted in a considerable improvement in reconstruction quality, whereas a negative reward penalizes weight configurations that decrease performance.

## 3.6 Objective Function

### 3.6.1 L1 Reconstruction Loss

To ensure basic integrity, a conventional pixel-wise $L1$ loss is applied between the student and ground truth.

$$\mathcal{L}_{rec} = \|\hat{I}_s - I_{GT}\| \tag{2}$$

This loss motivates the student to approximate the student's output quality in all regions.

### 3.6.2 Spatial-Weighted KD Loss

In our SPARK, we utilize the weight map $W$ to calculate the Spatial-weighted KD Loss, which is the weighted mean squared error (MSE) between the student output $\hat{I}_s$) and the teacher output $\hat{I}_t$).

$$\mathcal{L}_{KD} = \frac{1}{N} \sum_{i=1}^{N} W \odot (\hat{I}_s - \hat{I}_t)^2 \tag{3}$$

In here, $I_s$ and $I_t$ represent student and teacher outputs, respectively. $W$ is the spatially adaptive weight map predicted by the RL policy network, $N$ is the total number of spatial elements, and $\odot$ represents element-wise multiplication. This formulation prioritizes distillation intensity for challenging regions such as edges, textures, and high-loss zones, while smooth and well-reconstructed parts contribute less to the output.

#### 3.6.3 Total Training Objective

The final training objective incorporates the reconstruction and spatial-weighted KD losses, as shown below:

$$\mathsf{L}_{Total} = \mathsf{L}_{rec} + \mathsf{L}_{KD} \tag{4}$$

In here, the joint objective prioritizes pixel-level quality and spatially adaptable information transmission from teacher to student, focusing distillation pressure on perceptually difficult regions such as edges, fine textures, and high-error zones. Dynamically weighting each spatial region during training helps the quantized student model better allocate its limited representational capacity, rather than treating all regions uniformly. Targeted supervision allows students to attain reconstruction quality similar to the teacher, despite considerable computing constraints.

## 4 Experimental Setup and Results Analysis

### 4.1 Datasets and Methods

We evaluate our SPARK on three standard IR benchmarks. For LLE, we use the LOLv1 [15] dataset, which contains 485 training images and 15 test images captured under diverse low-light conditions. For SR, we evaluate on Urban100[5], a benchmark comprising 100 HR RGB images of urban scenes, including buildings, streets, architectural structures, windows, and repetitive structural patterns. For image denoising, we adopt the SIDD [1] dataset, which consists of 30,000 noisy–clean image pairs for training and 1,280 pairs for testing.

To evaluate our approach against the SOTA, we conducted experiments under four distinct configurations. In the first configuration, MIRNet [19] is employed as the teacher network, while an INT8-quantized variant of MIRNet serves as the student network. In the second configuration, MIRNet is utilized as the teacher network and the INT8-quantized variant of DnCNN [20] is utilized as the student network. For the SR task, two additional configurations are considered. In the third configuration, RFDN[6] is employed as the teacher network, while an INT8-quantized variant of RFDN serves as the student network. In the fourth configuration, RFDN is utilized as the teacher network and the INT8-quantized variant of SRCNN [2] is utilized as the student network.

### 4.2 Implementation Details

The student network is trained from scratch using L1 loss while we utilize the frozen teacher. The Student network is optimized using the Adam optimizer, with an initial learning rate set to 1e4. For training, input patches of size 128 × 128 with a batch size of 4. The entire framework is built upon PyTorch [11], and model quantization is carried out utilizing PyTorch quantization. All experiments were conducted on a single NVIDIA GeForce RTX 3080 GPU with 8 GB VRAM.

## 4.3 Results Analysis

### 4.3.1 Quantitative Evaluation

| Approach | Student Model | LLE (LOLv1)[15] | | | Denoising (SIDD) [1] | | |
|---|---|---|---|---|---|---|---|
| | | Avg PSNR↑ | Avg SSIM↑ | Avg LPIPS↓ | Avg PSNR↑ | Avg SSIM↑ | Avg LPIPS↓ |
| Full precision (teacher) | None | 24.14 | 0.801 | 0.150 | 39.72 | 0.959 | 0.325 |
| PTQ | | 20.10 (−1.91) | 0.751 (−0.034) | 0.194 (+0.018) | 28.42 (−5.47) | 0.510 (−0.297) | 0.577 (+0.257) |
| QAT | Self Quantized MIRNet [19] | 20.32 (−1.69) | 0.762 (−0.023) | 0.185 (+0.009) | 29.15 (−4.74) | 0.555 (−0.252) | 0.539 (+0.219) |
| KD [3] + QAT | | 21.45 (−0.56) | 0.778 (−0.007) | 0.179 (+0.003) | 29.57 (−4.32) | 0.571 (−0.236) | 0.527 (+0.207) |
| FAKD [18] + QAT | | 20.20 (−1.81) | 0.760 (−0.025) | 0.189 (+0.013) | 31.80 (−2.09) | 0.687 (−0.120) | 0.399 (+0.079) |
| DCKD [24] + QAT | | 19.92 (−2.09) | 0.751 (−0.034) | 0.192 (+0.016) | 31.72 (−2.17) | 0.680 (−0.127) | 0.402 (+0.082) |
| SLKD [23] + QAT | | 21.67 (−0.34) | 0.780 (−0.005) | 0.178 (+0.002) | 32.14 (−1.75) | 0.700 (−0.107) | 0.389 (+0.069) |
| **Our SPARK** | | **22.01** | **0.785** | **0.176** | **33.89** | **0.807** | **0.320** |
| PTQ | | 18.47 (−2.85) | 0.719 (−0.059) | 0.261 (+0.080) | 28.12 (−3.59) | 0.508 (−0.169) | 0.590 (+0.191) |
| QAT | Quantized DnCNN [20] | 19.01 (−2.31) | 0.720 (−0.058) | 0.211 (+0.030) | 28.45 (−3.26) | 0.520 (−0.157) | 0.579 (+0.180) |
| KD [3] + QAT | | 19.45 (−1.87) | 0.742 (−0.036) | 0.204 (+0.023) | 28.87 (−2.84) | 0.542 (−0.135) | 0.550 (+0.151) |
| FAKD [18] + QAT | | 20.88 (−0.44) | 0.761 (−0.017) | 0.191 (+0.010) | 29.36 (−2.35) | 0.558 (−0.119) | 0.541 (+0.142) |
| DCKD [24] + QAT | | 19.84 (−1.48) | 0.751 (−0.027) | 0.198 (+0.017) | 29.25 (−2.46) | 0.550 (−0.127) | 0.556 (+0.157) |
| SLKD [23] + QAT | | 20.97 (−0.35) | 0.764 (−0.014) | 0.190 (+0.009) | 29.34 (−2.37) | 0.563 (−0.114) | 0.530 (+0.131) |
| **Our SPARK** | | **21.32** | **0.778** | **0.181** | **31.71** | **0.677** | **0.399** |

Table 1: Comparison of quantization methods on LLE (LOLv1) and Denoising (SIDD). Green values show the difference relative to our approach. The teacher model is MIRNet [19] (full-precision), and the students are self-quantized MIRNet and quantized DnCNN [20].

Table 1 compares SPARK to PTQ, QAT, KD[3] + QAT, FAKD[18] + QAT, DCKD[24] + QAT and SLKD [23] + QAT for LLE (LOLv1) and Denoising (SIDD) using self-quantized MIRNet [19] and quantized DnCNN students. With 22.01 dB PSNR, 0.785 SSIM, and 0.176 LPIPS on LOLv1 and 33.89 dB PSNR, 0.807 SSIM, and 0.320 LPIPS [21] on SIDD, SPARK consistently outperforms all SOTA approaches under the self-quantized MIRNet student. Although KD + QAT, DCKD+QAT and FAKD + QAT show decent recoveries under low-bit limitations, the collapse of PTQ (-5.47 dB on SIDD) highlights the fragility of quantization-unaware approaches for restoration tasks.

Table 2 compares SPARK's performance on the SR task at ×2 and ×4 scales on Urban100 with self-quantized RFDN and quantized SRCNN students. At ×2, SPARK exceeds PTQ by +1.91 dB and achieves 32.01 dB PSNR, 0.825 SSIM, and 0.191 LPIPS under RFDN, decreasing the difference with the full-precision teacher (32.17 dB) to around -0.16 dB. At ×4, SPARK outperforms PTQ by +2.53 dB for RFDN and +1.22 dB for SRCNN. SPARK consistently achieves the highest PSNR, SSIM, and lowest LPIPS scores across scales and student models, indicating the capacity to properly restore full-precision teacher fidelity under actual compression constraints.

| Scale | Approach | Self-Quantized RFDN [6] (Student) | | | Quantized SRCNN [2] (Student) | | |
|---|---|---|---|---|---|---|---|
| | | Avg PSNR↑ | Avg SSIM↑ | Avg LPIPS↓ | Avg PSNR↑ | Avg SSIM↑ | Avg LPIPS↓ |
| | Full-precision (Teacher) | 32.17 | 0.928 | 0.178 | | — | |
| x2 | PTQ | 30.10 (−1.91) | 0.725 (−0.100) | 0.257 (+0.066) | 27.11 (−1.54) | 0.851 (−0.033) | 0.321 (+0.056) |
| | QAT | 31.04 (−0.97) | 0.738 (−0.087) | 0.244 (+0.053) | 27.85 (−0.80) | 0.868 (−0.016) | 0.310 (+0.045) |
| | KD [3] + QAT | 31.33 (−0.68) | 0.772 (−0.053) | 0.229 (+0.038) | 28.14 (−0.51) | 0.872 (−0.012) | 0.286 (+0.021) |
| | FAKD [18] + QAT | 31.45 (−0.56) | 0.805 (−0.020) | 0.216 (+0.025) | 28.20 (−0.45) | 0.879 (−0.005) | 0.278 (+0.013) |
| | DCKD [24] + QAT | 31.11 (−0.90) | 0.796 (−0.029) | 0.228 (+0.037) | 28.00 (−0.65) | 0.871 (−0.013) | 0.282 (+0.017) |
| | SLKD [23] + QAT | 31.57 (−0.44) | 0.809 (−0.016) | 0.216 (+0.025) | 28.28 (−0.37) | 0.879 (−0.005) | 0.277 (+0.012) |
| | **Our SPARK** | **32.01** | **0.825** | **0.191** | **28.65** | **0.884** | **0.265** |
| | Full-precision (Teacher) | 26.11 | 0.785 | 0.312 | | — | |
| x4 | PTQ | 23.36 (−2.53) | 0.654 (−0.065) | 0.426 (+0.032) | 22.56 (−1.22) | 0.690 (−0.030) | 0.489 (+0.071) |
| | QAT | 24.01 (−1.88) | 0.681 (−0.038) | 0.415 (+0.021) | 22.70 (−1.08) | 0.693 (−0.027) | 0.458 (+0.040) |
| | KD [3] + QAT | 24.56 (−1.33) | 0.699 (−0.020) | 0.411 (+0.017) | 23.32 (−0.46) | 0.711 (−0.009) | 0.445 (+0.027) |
| | FAKD [18] + QAT | 25.07 (−0.82) | 0.712 (−0.007) | 0.405 (+0.011) | 23.45 (−0.33) | 0.715 (−0.005) | 0.430 (+0.012) |
| | DCKD [24] + QAT | 25.09 (−0.80) | 0.710 (−0.009) | 0.414 (+0.020) | 23.52 (−0.26) | 0.719 (−0.001) | 0.430 (+0.012) |
| | SLKD [23] + QAT | 25.19 (−0.70) | 0.712 (−0.007) | 0.401 (+0.007) | 23.54 (−0.24) | 0.715 (−0.005) | 0.429 (+0.011) |
| | **Our SPARK** | **25.89** | **0.719** | **0.394** | **23.78** | **0.720** | **0.418** |

Table 2: Comparison of quantization methods at ×2 and ×4 scales on the Urban100 dataset. Green values show the difference relative to our approach. The teacher model is RFDN (full-precision), and the students are self-quantized RFDN and quantized SRCNN.

### 4.3.2 Qualitative Evaluation

Fig. 3 demonstrates a qualitative comparison of SPARK to compete with KD approaches, notably FAKD, DCKD, and SLKD, on three IR tasks. FAKD, DCKD, and SLKD all fail to recover brightness and structural detail from highly underexposed inputs, but our SPARK recovers scene content and contrast most accurately to the ground truth. In the denoising, other approaches show residual noise artifacts and color distortion in smooth regions, but SPARK offers cleaner results with higher color uniformity and sharper edge boundaries. On the other hand, SOTA approaches for ×4 SR suffer from over-smoothing and aliasing in high-frequency regions such as diagonal architectural edges. However, SPARK reconstructs finer structural features and more accurate geometry. Across all three tasks, the proposed SPARK consistently produces perceptually better outcomes, demonstrating its ability to transmit complex representational knowledge from teacher to student networks.

## 4.4 Ablation Study

To further verify the effectiveness of our RL-based KD approach, we conducted ablation studies comparing three settings, namely training with RL, training without RL using L1 loss, and training without RL using MSE loss.

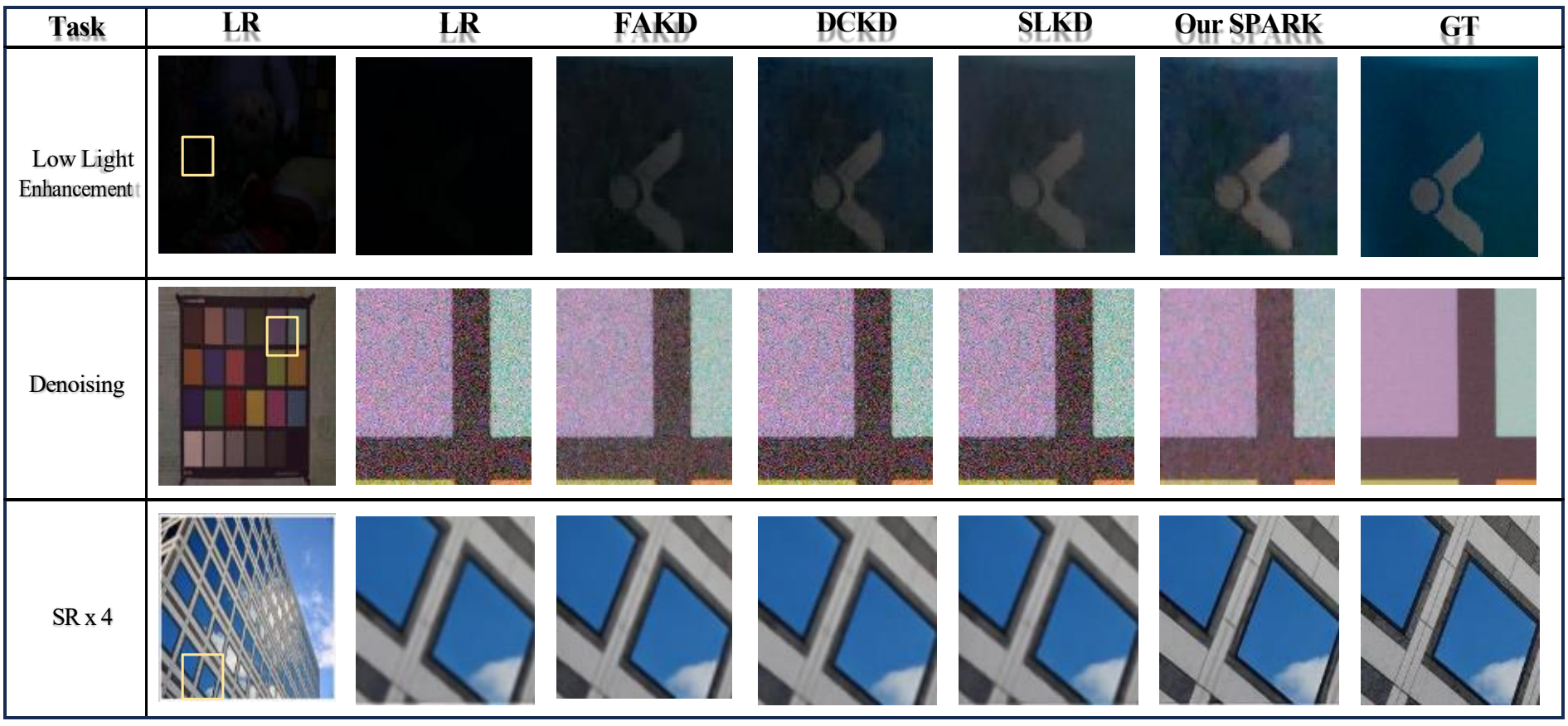

Figure 3: Qualitative comparison of SPARK against SOTA KD methods (FAKD, DCKD, SLKD) on three IR tasks: low-light enhancement(LOLv1), image denoising(SIDD), and SR (×4 on Urban100). Ground truth (GT) is shown in the last column for reference.

Table 3: Ablation study on the effect of RL across two student models on LLE (LOLv1) and Denoising (SIDD) tasks. Best results are shown in **bold**.

| Approach | Student Model | LLE (LOLv1) | | | Denoising (SIDD) | | |
|---|---|---|---|---|---|---|---|
| | | Avg. PSNR | Avg. SSIM | Avg. LPIPS | Avg. PSNR | Avg. SSIM | Avg. LPIPS |
| With RL | Self Quantized MIRNet[19] | **22.01** | **0.785** | **0.176** | **33.89** | **0.807** | **0.320** |
| Without RL (L1 loss) | | 21.45 | 0.778 | 0.179 | 29.57 | 0.571 | 0.527 |
| Without RL (MSE loss) | | 21.48 | 0.778 | 0.185 | 29.63 | 0.572 | 0.531 |
| With RL | Quantized DnCNN [20] | **21.32** | **0.778** | **0.181** | **31.71** | **0.677** | **0.399** |
| Without RL (L1 loss) | | 19.45 | 0.742 | 0.204 | 28.87 | 0.542 | 0.550 |
| Without RL (MSE loss) | | 19.46 | 0.749 | 0.218 | 28.91 | 0.554 | 0.567 |

Tables 3 and 4 demonstrate that the RL setting consistently outperforms all student models, tasks, and SR scales in terms of PSNR, SSIM, and LPIPS measures. These results demonstrate that the RL-based reward mechanism plays an important role in steering the student network toward higher-quality reconstruction as compared to traditional loss functions.

# 5 Discussion & Limitations

Our proposed SPARK consistently outperforms uniform KD baselines across many challenges and student architectures, validating the core hypothesis that spatially adaptive distillation is more effective than treating all image areas similarly under low-bit quantization limitations. The interpretable difficulty features offer a rational foundation for the policy's decisions, while stochastic Gaussian sampling guarantees enough exploration during training to prevent trivial or degenerate weight combinations. However, the limitations of this approach is the reward signal is tied to global reconstruction metrics such as PSNR and SSIM, which may not fully capture perceptual quality in all degradation scenarios. Addressing this limitation is an important direction for future work.

Table 4: Ablation study on the effect of RL for SR across two student models at scales ×2 and ×4 for Urban100 dataset. Best results are shown in **bold**.

| Scale | Approach | Self-Quantized RFDN as Student | | | Quantized SRCNN as Student | | |
|---|---|---|---|---|---|---|---|
| | | Avg. PSNR | Avg. SSIM | Avg. LPIPS | Avg. PSNR | Avg. SSIM | Avg. LPIPS |
| ×2 | With RL | **32.01** | **0.825** | **0.191** | **28.65** | **0.884** | **0.265** |
| | Without RL (L1 loss) | 31.33 | 0.772 | 0.229 | 28.14 | 0.872 | 0.286 |
| | Without RL (MSE loss) | 31.41 | 0.780 | 0.230 | 28.16 | 0.876 | 0.291 |
| ×4 | With RL | **25.89** | **0.709** | **0.394** | **23.78** | **0.720** | **0.418** |
| | Without RL (L1 loss) | 24.56 | 0.699 | 0.411 | 23.32 | 0.711 | 0.445 |
| | Without RL (MSE loss) | 24.58 | 0.701 | 0.422 | 23.40 | 0.711 | 0.457 |

# Conclusion

In this study, we propose SPARK, a spatial policy-driven adaptive RL framework for KD in low-bit IR. In contrast to existing KD approaches, which uniformly supervise all spatial regions, SPARK utilizes a lightweight RL policy network to dynamically produce a spatially adaptive weight map based on interpretable difficulty features, concentrating distillation effort on perceptually challenging regions such as edges, textures, and higher-error zones. The policy is learned via reward signals associated with reconstruction quality, allowing for exploration of non-trivial spatial weight combinations that simply differentiable objectives cannot achieve. Extensive studies with LLE, image denoising, and SR indicate that SPARK consistently outperforms SOTA KD approaches across a variety of student architectures and compression settings. Ablation studies demonstrate that the RL-based reward mechanism is the primary cause of improvement above traditional L1 and MSE loss alternatives. Our proposed SPARK establishes a foundation for spatially adaptive distillation in low-level vision and opens up a new direction for RL-guided compression of IR networks.